\documentclass{article}

\usepackage[preprint]{neurips_2026}

\usepackage[utf8]{inputenc}
\usepackage[T1]{fontenc}
\usepackage{hyperref}
\usepackage{url}
\usepackage{booktabs}
\usepackage{amsfonts}
\usepackage{nicefrac}
\usepackage{microtype}
\usepackage{xcolor}
\usepackage{multirow}
\usepackage{rotating}
\usepackage{colortbl}
\newcommand{\best}[1]{\textbf{#1}}
\newcommand{\second}[1]{\underline{#1}}
\usepackage[most]{tcolorbox}
\usepackage{graphicx}
\definecolor{oursgray}{gray}{0.92}
\newcommand{\ourscell}[1]{\cellcolor{oursgray}{#1}}
\title{Object-Uni: A Unified Model for Object-Centric Spatial Understanding and Controllable Generation}

\author{
Mining Tan$^{1,2 \dagger}$ \quad Yinuo Wang$^{3 \dagger}$ \quad Ziqi Zhou$^{2,1}$ \quad Weize Quan$^{2,1}$ \\
\textbf{Sifei Li$^{2,1}$ \quad Jingdong Chen$^4$ \quad
DanDan Zheng$^4$ \quad Libin Wang$^4$ \quad Weiming Dong$^{2,1}$} \thanks{Corresponding author <weiming.dong@ia.ac.cn>.} \vspace{0.15cm}\\
$^1$School of Artificial Intelligence, University of Chinese Academy of Sciences \\ $^2$MAIS, Institute of Automation, Chinese Academy of Sciences \\ $^3$School of Vehicle and Mobility, Tsinghua University \\ $^4$Ant Group, China
}

\begin{document}

\maketitle

\begin{abstract}
Unified models for visual understanding and generation have made rapid progress, yet they still lack the ability to understand and manipulate the spatial states of object instances. Existing models can describe objects in natural language, but they struggle to precisely represent continuous object poses and generate geometrically consistent images under target viewpoints. To mitigate this, we propose \emph{Object-Uni}, a unified model for object-centric spatial understanding and controllable generation. Specifically, we formulate object-centric spatial intelligence as a unified problem connecting pose perception, spatial reasoning, pose-conditioned generation, and object-centric novel view synthesis. We treat object pose as an explicit geometric variable shared by understanding and generation, rather than merely a prediction label or control signal. To make pose usable by multimodal large language models, we propose a viewpoint-based orientation abstraction that maps orientation into structured viewpoint descriptions while preserving continuous geometric supervision. We further construct an object-centric spatial benchmark (UniSpatial-80K) and train a unified model with an object-token-grounded pose anchor to associate each instance with its pose state. Experiments show that our model improves object-level pose understanding and pose-controllable generation, moving unified models from describing objects toward manipulating spatial states.
\end{abstract}

\section{Introduction}
The rapid development of unified models for visual understanding and generation is pushing visual intelligence from merely “understanding images” toward “understanding and generating the visual world.” Recent unified models~\citep{chameleon2024, zhou2025transfusion, xie2025showo, deng2025emerging, wu2025janus, ai2025ming} have demonstrated strong capabilities in image captioning, visual question answering, image generation, and image editing, showing impressive progress in open-ended semantic understanding and high-quality visual synthesis. However, real-world visual intelligence requires more than recognizing objects and semantic content in images. A model should also understand the spatial location and orientation of object instances, as well as how an object should appear when the viewpoint changes. We refer to this capability as \textbf{object-centric spatial intelligence}: the ability to reason about object instances in terms of their spatial locations, poses, and relative relations, and to support pose-conditioned controllable generation and object-centric novel view synthesis.

Current methods typically rely on free-form textual descriptions to represent object spatial states, such as “the shoe on the left,” “a car viewed from a high angle,” or “a chair facing right.” \citep{nichols2025right, ma2025spatialllm} Such descriptions are natural and flexible, but they lack stable geometric constraints and are insufficient for precisely representing continuous object poses. Meanwhile, conventional object pose estimation mainly focuses on predicting numerical pose values from images, whereas pose-conditioned generation aims to synthesize target-view images under given pose conditions. These two lines of research are usually studied as separate tasks, lacking a unified problem formulation that jointly characterizes object pose perception, spatial reasoning, and pose-conditioned generation.

\begin{figure}
    \centering
    \includegraphics[width=0.95\linewidth]{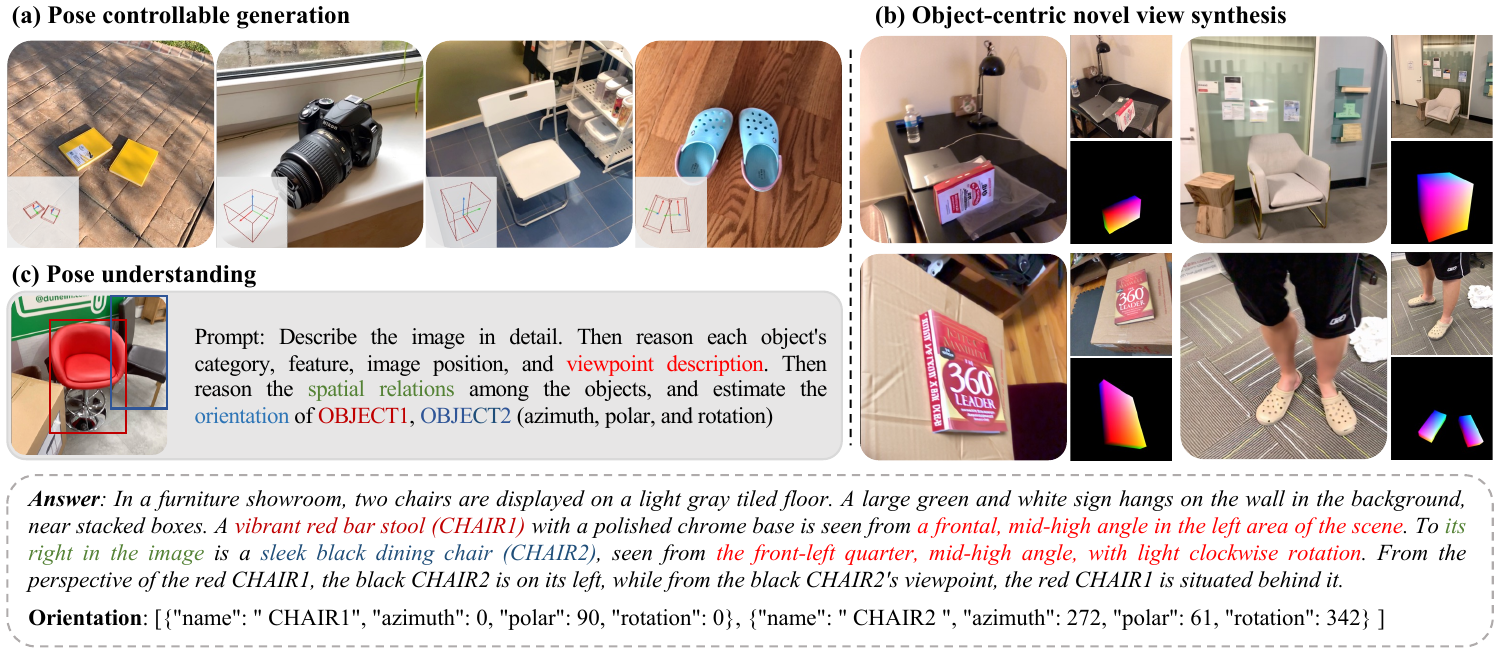}
    \caption{Versatile capabilities of Object-Uni model. It unifies (a) object-centric generation and (c) understanding, also supporting orientation estimation and (b) object-centric novel view synthesis.}
    \label{fig:ability}
    \vspace{-2mm}
\end{figure}

In this work, we formulate object-centric spatial intelligence as a unified understanding-and-generation problem. Under this formulation, a model is required not only to infer the spatial state of objects from images but also to generate geometrically consistent visual content according to a target spatial state. Our central insight is that object pose should not be treated merely as a label for pose estimation, nor only as an external control signal for image generation. Instead, it should serve as an explicit object-centric geometric variable that unifies pose perception, spatial reasoning, and pose-conditioned generation. The key challenge is how to make object pose understandable and generable by multimodal large language models (MLLMs). Object locations and relative spatial relations can often be expressed naturally in language, such as left of, behind, or on top of. In contrast, object orientation is a multi-dimensional continuous geometric variable that is difficult to directly align with the linguistic reasoning space of MLLMs. To address this issue, we propose a viewpoint-based orientation abstraction, which maps azimuth, polar, and rotation into structured viewpoint descriptions, such as front view, high-angle view, and clockwise rotation. This abstraction does not replace continuous pose with discrete viewpoint labels. Instead, it builds a bridge between continuous geometric supervision and language-level spatial reasoning: continuous pose preserves precise geometric constraints and enables quantitative evaluation, while structured viewpoint descriptions allow the model to understand and reason about object pose in natural language.

We further construct an object-centric spatial understanding and generation benchmark with 80K+ samples. The benchmark covers complex scenarios, including street-view vehicles and pedestrians, indoor objects, and general indoor and outdoor objects. For each object instance, it organizes spatial location, relative relations, continuous pose, and viewpoint-based descriptions. Unlike conventional pose estimation datasets that focus on a single task, our benchmark is organized into a set of interconnected understanding and generation tasks, including object spatial-relation understanding, viewpoint reasoning, orientation prediction, pose-conditioned image generation, and object-centric novel view synthesis. These tasks forms a unified evaluation protocol: a model can both understand object poses from images and generate geometrically consistent images according to target poses.

Based on this benchmark, we train a unified object-centric spatial intelligence model that performs object pose understanding, viewpoint reasoning, and pose-controllable generation within a single framework. To enable stable orientation understanding in both single-object and multi-object scenes, we design an object-token-grounded pose anchor that aggregates visual features corresponding to each object instance and explicitly associates each object with its orientation state. We illustrate the versatile capabilities of our Object-Uni model in Figure~\ref{fig:ability}. Our contribution can be summarized as:

\begin{itemize}
\item We define object-centric spatial intelligence as a unified understanding-and-generation problem. Unlike prior studies that treat object pose estimation and pose-conditioned generation as separate tasks, we model object pose as an explicit object-centric geometric variable that connects pose perception, spatial reasoning, and controllable generation.

\item We propose a viewpoint-based orientation abstraction, which converts continuous 3D orientations into structured natural-language viewpoint descriptions. It allows MLLMs to understand and reason about object orientation through natural viewpoint terms.

\item We construct an object-centric spatial understanding-and-generation benchmark with 80K+ samples. The benchmark covers both single-object and multi-object scenarios and unifies object spatial-relation understanding, viewpoint reasoning, orientation prediction, and pose-conditioned generation under systematic evaluation.

\item We train a unified model for object-centric spatial intelligence through multi-stage training, achieving superior performance on both understanding and generation tasks. In addition, for orientation prediction during the understanding stage, we introduce an object-token-grounded pose anchor that explicitly associates each predicted orientation with its corresponding object instance, thereby improving prediction performance in multi-object scenes.
\end{itemize}

\section{Related Work}
\paragraph{Object Pose Understanding from Images}
Understanding object pose and orientation is a fundamental requirement for object-centric spatial intelligence. Early object-centric 3D benchmarks such as Objectron \citep{ahmadyan2021objectron} provide videos with camera poses, sparse point clouds, and manually annotated 3D bounding boxes describing object position, orientation, and dimensions, while Omni3D \citep{brazil2023omni3d} studies large-scale 3D object detection across diverse indoor and outdoor scenes. More recently, Orient Anything \citep{wang2024orient} formulates open-world object orientation estimation from a single image by rendering large-scale 3D assets with precise azimuth, polar, and in-plane rotation annotations. Orient Anything V2 \citep{wang2026orient} further extends this formulation to rotationally symmetric objects and relative rotation estimation. In parallel, recent MLLM-oriented benchmarks such as Right Side Up \citep{nichols2025right} reveal that current multimodal models still struggle with fine-grained multi-axis orientation understanding. Unlike these works, which primarily evaluate or predict object pose, our goal is to make object orientation a shared variable for both multimodal reasoning and controllable generation.

\paragraph{Spatially Controlled Generation}
Controlling object pose or viewpoint in text-to-image generation has been explored from several directions. Continuous 3D Words \citep{cheng2024learning} encode continuous 3D-aware attributes, including object pose, as special tokens that can be smoothly varied during generation. Custom Diffusion 360 \citep{kumari2024customizing} focuses on subject-driven generation with viewpoint control for customized objects from multi-view images. Compass Control \citep{parihar2025compass} addresses multi-object orientation control by introducing object-specific compass tokens and constraining their attention maps to corresponding object regions. SceneDesigner \citep{ribeiro2021scene} further studies controllable multi-object image generation with 9-DoF pose manipulation and uses CNOCS maps as dense geometric pose conditions. The main difference between our generation module with those works is that we do not treat pose solely as a generation control signal; instead, the same object-level pose representation is also predicted and reasoned about by the MLLM side, enabling a unified understanding-and-generation framework.

\paragraph{Unified Models}
Recent unified multimodal models aim to support visual understanding and generation within a single framework. Chameleon \citep{team2024chameleon}, Emu2 \citep{sun2024generative}, Lumina-mGPT \citep{liu2024lumina}, Show-O \citep{xie2024show}, and Janus \citep{wu2025janus} explore different ways to unify image-text comprehension and generation, such as mixed-modal autoregressive modeling, autoregressive-diffusion hybrid objectives, or decoupled visual encoders. These works mainly focus on architecture- or objective-level unification for general multimodal tasks. In contrast, our work focuses on object-centric spatial unification: we make object orientation a shared variable that can be described, predicted, grounded to an object token, and reused as a control signal for image generation. Therefore, Object-Uni is not intended as a new general-purpose unified architecture, but as a pose-centered interface built on a unified multimodal backbone for bridging object-level spatial understanding and controllable generation.

\begin{figure}
    \centering
    \includegraphics[width=1.0\linewidth]{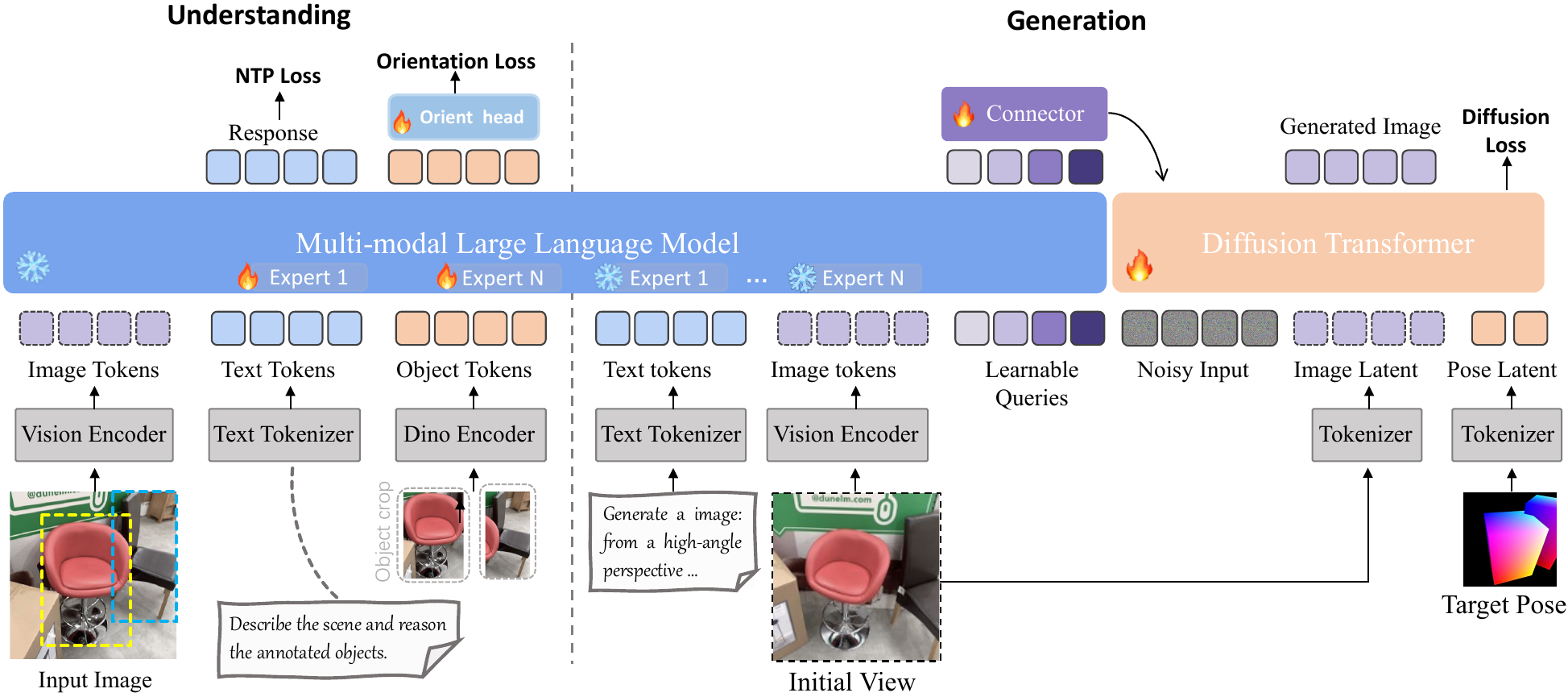}
    \caption{Overview of Object-Uni. Our model unifies object-centric spatial understanding and orientation-controllable generation. The dotted modules denote object-centric novel view synthesis.}
    \label{fig:arc}
\end{figure}

\section{Object-Centric Understanding and Controllable Generation}

\subsection{Overview}
We study object-centric spatial intelligence in a unified understanding-and-generation framework. As shown in Figure~\ref{fig:arc}, given an image, a text prompt, and object-level spatial conditions, the model is required to infer object locations, relative spatial relations, continuous orientations, and further generate images that satisfy the specified object-centric pose conditions.
The main difficulty lies in the representational mismatch between object pose and multimodal language modeling.
Object pose is a continuous and periodic geometric variable, while MLLMs use discrete tokens and natural-language descriptions.
To bridge this gap, we introduce a viewpoint-based orientation abstraction that maps continuous orientation into structured spatial language while preserving the original continuous pose for metric supervision and controllable generation.
Based on this, we construct the UniSpatial-80K benchmark, and train an object-centric unified multimodal model for pose understanding, spatial reasoning, pose-controllable generation, and object-centric novel view synthesis.

\subsection{Viewpoint-based Orientation Abstraction}

For each object instance $\mathrm{obj}_i$ in an image $I$, we denote its 2D bounding box as
$b_i = (x_i^1, y_i^1, x_i^2, y_i^2)$ and its 9D pose as
$q_i = (\mathbf{l}_i, \mathbf{s}_i, \mathbf{o}_i)$.
Here, $\mathbf{l}_i \in \mathbb{R}^3$ denotes the 3D location of the object center in the camera coordinate system,
$\mathbf{s}_i \in \mathbb{R}^3$ is its 3D size, and
$\mathbf{o}_i = (\phi_i, \theta_i, \psi_i)$ denotes its 3D orientation from the camera view, where $\phi_i \in [0^\circ, 360^\circ)$ is the azimuth,
$\theta_i \in [0^\circ, 180^\circ]$ is the polar, and
$\psi_i \in [0^\circ, 360^\circ)$ is the rotation angle. We follow the orientation convention of Orient Anything~\citep{wang2025orient} for defining these three angles.
The azimuth describes the object-facing direction relative to the camera, the polar angle is the camera elevation with respect to the object, and the rotation angle describes the in-plane image rotation.
This continuous representation is necessary for quantitative pose supervision and evaluation.
However, directly exposing raw angles is hard for MLLM to understand, since numerical tokens do not naturally encode the perceptual meaning of object viewpoint.
We therefore construct a dual representation of object pose: the continuous angles remain as metric targets, while a deterministic language abstraction converts them into viewpoint descriptions.

To bridge continuous object orientation and natural language descriptions, we map the three orientation angles
$\mathbf{o}_i=(\phi_i,\theta_i,\psi_i)$ into coarse semantic view terms.
For azimuth, we divide $\phi_i$ into eight canonical object views, such as front, front-right quarter, and right side. For polar angle, we map $\theta_i$ into camera-elevation descriptions.
For in-plane rotation, we simply divide it into three intervals. The detailed mapping rules are summarized in Appendix \ref{app: view_mapp}.

In addition to orientation, we describe each object using its coarse image-space location and pairwise spatial relations.
The normalized bounding-box center $(\bar{x}_i,\bar{y}_i)$ is mapped to horizontal and vertical region terms, such as left, center, right, top, middle, and bottom.
For multi-object scenes, we describe relations from both viewer-centric and object-centric perspectives.
The former is computed from relative image positions, while the latter further incorporates the azimuth of the reference object to express relations such as left of, right of, in front of, or behind in the object's local frame.
This provides language supervision that jointly reflects image layout and object-centric geometry.

\subsection{UniSpatial-80K Dataset}

\paragraph{Data Sources}
We construct UniSpatial-80K, a large-scale object-centric spatial understanding-and-generation benchmark.
The benchmark integrates multiple object-level 3D and pose-annotated data sources, including the KITTI \citep{geiger2012we}, Cityscapes \citep{cordts2016cityscapes} and Objectron \citep{ahmadyan2021objectron} subsets from OmniNOCS \citep{krishnan2024omninocs}, and ImageNet3D \citep{ma2024imagenet3d}.
These sources provide complementary scene distributions.
KITTI and Cityscapes provide street-view data containing realistic outdoor layouts with vehicles and pedestrians.
Objectron provides object-centric indoor videos with background context.
ImageNet3D provides broad category coverage and general-purpose object-level 3D annotations.
Together, they allow us to evaluate object-centric spatial intelligence beyond a narrow category or scene type.

\paragraph{Data Filtering and Balancing}
We filter the raw data to ensure reliable object-centric spatial supervision.
First, we remove samples with excessive background clutter or too many foreground instances, as these cases make object–pose attribution ambiguous. Second, we exclude categories or samples whose canonical front direction is not well defined, because orientation supervision becomes ill posed when an object lacks a unique or stable front face. Third, to mitigate severe category imbalance, we cap overrepresented categories by sampling at most 5,000 images per category, to improve the reliability of continuous orientation labels and makes the benchmark more suitable for both understanding and generation tasks. UniSpatial-80K contains 83,252 image entries with 91,392 annotated objects across 122 training categories, the detailed distribution and statistics are provided in Appendix~\ref{app_data}.

\paragraph{Pose Conversion and Spatial Caption Construction}
For each retained object instance, we convert the original dataset-specific pose annotations into the unified convention.
We then apply the viewpoint-based orientation abstraction described above to obtain structured spatial descriptions.
Each final sample contains a general image caption, an object-centric spatial caption, and continuous pose annotations. The general caption describes the visual content and scene context, while the spatial caption explicitly describes object locations, viewpoints, rotations, and inter-object relations. The example is showned in the Figure~\ref{fig:ability} (c).

\subsection{Object-Centric Unified Multimodal Model}

\paragraph{Base Architecture}
We build our model upon Ming-Lite-Uni\citep{ai2025ming}, a unified multimodal architecture that connects a multimodal autoregressive model with diffusion transformer blocks through multi-scale learnable query representations.
The MLLM side receives text instructions and visual tokens, performs multimodal understanding and spatial reasoning, and produces hidden states for language generation.
The generation side uses a DiT-based diffusion model to synthesize or edit images conditioned on text, visual features, and spatial control signals.
This architecture provides a natural basis for unifying object-centric understanding and controllable generation: the MLLM handles semantic and spatial reasoning, while the diffusion model handles high-fidelity image synthesis. The overall architecture is shown in Figure~\ref{fig:arc}.

\paragraph{Object-Token-Grounded Pose Anchor}
A direct way to predict orientation is to ask the MLLM to output numerical angles in text and then parse the generated values.
However, this strategy is unstable, since numerical angle tokens are weakly calibrated, and in multi-object scenes, the generated text may fail to associate each angle with the correct object instance.
To address this issue, we introduce an object-token-grounded pose anchor, PoseAnchor.
For each object $o_i$, we crop the object region according to its bounding box and extract an object-level feature using a frozen vision encoder \citep{oquab2023dinov2}.
The object feature is projected into the MLLM hidden space and inserted into the input sequence as an object-specific token.
The hidden state corresponding to this token serves as an instance-level pose anchor:
\begin{equation}
h_i = \mathrm{MLLM}(I, T, b_i, e_i),
\end{equation}
where $e_i$ denotes the projected object feature.
We then attach lightweight prediction heads to $h_i$ to estimate azimuth, polar, and rotation. This design explicitly binds each pose prediction to the corresponding object instance, which is especially important in multi-object scenes. 

\paragraph{Pose Prediction Loss}
Following Orient Anything~\cite{wang2025orient}, we formulate object orientation prediction as a distribution fitting problem over discretized angle bins. For each valid object, the model predicts three distributions corresponding to azimuth $\phi$, polar $\theta$, and rotation $\psi$. The soft target distribution is constructed by placing a Gaussian-like distribution around the ground-truth angle. For azimuth and rotation, we compute the target using circular angular distance to respect periodicity, while for the polar angle we use a non-periodic distance on $[0, 180^\circ]$.

Let $p_i^\phi$, $p_i^\theta$, and $p_i^\psi$ denote the predicted distributions, and let
$q_i^\phi$, $q_i^\theta$, and $q_i^\psi$ denote the corresponding soft target distributions.
The pose loss is defined as the average soft cross-entropy over valid objects:
\begin{equation}
\mathcal{L}_{\mathrm{pose}}
=
\frac{1}{|\mathcal{V}|}
\sum_{i \in \mathcal{V}}
\left[
H(q_i^\phi, p_i^\phi)
+
H(q_i^\theta, p_i^\theta)
+
H(q_i^\psi, p_i^\psi)
\right],
\end{equation}
where $H(q,p)=-\sum_k q_k \log p_k$. The full understanding-stage objective is:
\begin{equation}
\mathcal{L}_{\mathrm{und}}
=
\mathcal{L}_{\mathrm{NTP}}
+
\lambda_{\mathrm{pose}}\mathcal{L}_{\mathrm{pose}}.
\end{equation}

\paragraph{Task Formulation}
UniSpatial-80K is organized as a set of interconnected spatial understanding and generation tasks rather than as a single pose-estimation benchmark. For object-centric spatial understanding, we define the \textit{Image} $\rightarrow$ \textit{Spatial Text + Orient} task, in which the model generates a spatial description and predicts the continuous orientation of each object in the input image. For pose-controllable generation, we define the \textit{Text + Pose} $\rightarrow$ \textit{Image} task, where the model synthesizes an image consistent with both the textual prompt and the specified object pose conditions. To further leverage the spatial reasoning capabilities of MLLMs, we introduce the \textit{Text + Pose} $\rightarrow$ \textit{Spatial Text} task, in which the model converts textual content and pose conditions into structured spatial descriptions. This task serves as an implicit reasoning bridge between spatial understanding and generation. Finally, for object-centric novel view synthesis, we define the \textit{Image + Pose} $\rightarrow$ \textit{Image} task, where the model synthesizes a target-view image according to the specified target object pose.

Since language provides only coarse spatial constraints, we additionally encode the target object pose as a generation-side geometric condition. Following CNOCS-based 9-DoF pose control~\citep{qinscenedesigner}, we convert the object pose into a CNOCS map. The CNOCS map provides dense image-space geometric guidance and is encoded into the diffusion model through the VAE encoder.

\section{Experiments}
We conduct experiments to evaluate whether the proposed model can develop object-centric spatial capabilities for both understanding and generation. Specifically, we study three questions: 
(1) Can the unified model accurately estimate object orientation from images?
(2) Can it effectively control pose for text-to-image generation?
(3) Does PoseAnchor improve the orientation estimation, especially in multi-object scenes?

\subsection{Implementation Details}

In our implementation, the MLLM-side input image is encoded by a vision encoder, whereas the DiT-side reference image and CNOCS map are encoded by the VAE encoder. Each pose anchor \texttt{<obj>} is grounded by an object token, associated with a 2D bounding box, and inserted into the input sequence to predict the corresponding orientation tuple. Training proceeds in three stages. First, we train the MLLM-side spatial reasoning component with two understanding-oriented tasks: \textit{Image} $\rightarrow$ \textit{Spatial Text + Orient} and \textit{Text + Pose} $\rightarrow$ \textit{Spatial Text}. We update only the MLLM Mixture-of-Experts (MoE) layers, the object-token projection module, and the orientation prediction heads, while keeping the rest of the pretrained backbone frozen. Second, we train the DiT generator for orientation-controllable text-to-image generation, i.e., \textit{Text + Pose} $\rightarrow$ \textit{Image}. Finally, we train the model for object-centric novel view synthesis, i.e., \textit{Image + Pose} $\rightarrow$ \textit{Image}. The full training pipeline takes approximately 18 hours on 8 H20 GPUs.

\subsection{Pose Understanding}

\begin{table*}[t]
  \caption{\textbf{Evaluation results on object orientation estimation.} The results are grouped by dataset and split. Best results in each dataset-split block are shown in bold, and second-best results are underlined. The Object-Uni w/o represents our model without object-token-grounded pose anchor.}
  \label{tab:main_orientation}
  \centering
  \scriptsize
  \renewcommand{\arraystretch}{0.95}%
  \setlength{\tabcolsep}{2.4pt}%
  \resizebox{\textwidth}{!}{%
  \begin{tabular}{cc l cccc cccc cccc}
  \toprule
  \multirow{3}{*}{} & \multirow{3}{*}{} & \multirow{3}{*}{Approach} 
  & \multicolumn{4}{c}{Azimuth [in $^\circ$]} 
  & \multicolumn{4}{c}{Polar [in $^\circ$]} 
  & \multicolumn{4}{c}{Rotation [in $^\circ$]} \\
  \cmidrule(lr){4-7}\cmidrule(lr){8-11}\cmidrule(lr){12-15}
  &&& Error$\downarrow$ & \multicolumn{3}{c}{AUC@5/10/30$\uparrow$}
  & Error$\downarrow$ & \multicolumn{3}{c}{AUC@5/10/30$\uparrow$}
  & Error$\downarrow$ & \multicolumn{3}{c}{AUC@5/10/30$\uparrow$} \\
  \midrule
  
  \multirow{14}{*}{\begin{sideways}\textbf{KITTI-Cityscapes}\end{sideways}} 
  & \multirow{7}{*}{\begin{sideways}Single\end{sideways}} 
  & Random & 86.18 & 2.81 & 4.16 & 10.47 & 43.57 & 4.25 & 8.04 & 19.27 & 94.49 & 1.36 & 2.82 & 8.36 \\
  & & GPT-4o & 73.38 & 7.52 & 13.48 & 32.05 & 2.93 & 45.60 & 72.41 & 90.80 & 0.78 & 99.30 & 99.30 & 99.47 \\
  & & Gemini-2.5-pro & 69.32 & 10.14 & 16.21 & 31.78 & 2.92 & 47.87 & 72.64 & 90.88 & 0.80 & 99.26 & 99.63 & 99.88 \\
  & & OriAny.V1 & \second{27.33} & 14.99 & 28.03 & \second{56.72} & 7.61 & 23.42 & 49.55 & 79.24 & 1.53 & 98.29 & 98.73 & 99.03 \\
  & & OriAny.V2 & 29.94 & 15.48 & 27.07 & 55.40 & 2.34 & 59.72 & 79.44 & 93.09 & 3.65 & 31.46 & 65.61 & 88.54 \\
  & & Object-Uni w/o & 46.16 & \second{28.56} & \second{38.30} & 54.76 & \second{0.78} & \second{93.88} & \second{96.94} & \second{98.98} & \second{0.40} & \second{99.45} & \second{99.72} & \second{99.91} \\
  & & \ourscell{Object-Uni} & \best{22.87} & \best{41.54} & \best{56.82} & \best{74.44} & \best{0.56} & \best{98.18} & \best{99.09} & \best{99.70} & \best{0.23} & \best{99.67} & \best{99.83} & \best{99.94} \\
  \cmidrule(lr){2-15}
  
  & \multirow{7}{*}{\begin{sideways}Multi\end{sideways}} 
  & Random & 90.20 & 2.13 & 3.42 & 8.85 & 43.73 & 4.42 & 7.53 & 18.74 & 89.87 & 1.52 & 2.98 & 9.46 \\
  & & GPT-4o & 82.94 & 9.69 & 13.65 & 28.16 & 1.52 & 80.34 & 89.04 & 96.31 & 0.67 & 99.62 & 99.86 & 99.94 \\
  & & Gemini-2.5-pro & 48.43 & 24.52 & 31.18 & 42.96 & 2.64 & 58.30 & 76.75 & 92.14 & 0.68 & 99.62 & 99.81 & 99.94 \\
  & & OriAny.V1 & \best{22.17} & 22.57 & 40.04 & \second{69.04} & 10.37 & 49.83 & 66.15 & 81.65 & 3.27 & 91.70 & 94.53 & 96.42 \\
  & & OriAny.V2 & 54.93 & 17.03 & 24.12 & 48.33 & 0.99 & 90.39 & 95.04 & 98.31 & 4.35 & 14.62 & 57.15 & 85.70 \\
  & & Object-Uni w/o & 52.02 & \second{30.00} & \second{41.61} & 55.58 & \second{0.92} & \second{90.46} & \second{95.14} & \second{98.38} & \second{0.41} & \second{99.72} & \second{99.86} & \second{99.95} \\
  & & \ourscell{Object-Uni} & \second{28.12} & \best{47.06} & \best{60.32} & \best{74.04} & \best{0.46} & \best{97.05} & \best{98.48} & \best{99.49} & \best{0.19} & \best{99.75} & \best{99.87} & \best{99.96} \\
  \midrule
  
  \multirow{14}{*}{\begin{sideways}\textbf{ImageNet3D}\end{sideways}} 
  & \multirow{7}{*}{\begin{sideways}Single\end{sideways}} 
  & Random & 87.52 & 2.10 & 3.74 & 10.04 & 46.51 & 3.60 & 7.04 & 18.35 & 92.03 & 1.25 & 2.76 & 8.30 \\
  & & GPT-4o & 50.02 & 17.26 & 21.24 & 33.88 & 15.31 & 26.57 & 34.82 & 57.09 & 10.05 & 67.51 & 72.52 & 82.37 \\
  & & Gemini-2.5-pro & 53.77 & 11.18 & 14.24 & 28.29 & 13.14 & 22.95 & 36.08 & 64.65 & 10.79 & 65.09 & 71.25 & 81.07 \\
  & & OriAny.V1 & 41.75 & 14.03 & 24.21 & 46.35 & 26.32 & 11.28 & 18.52 & 39.52 & 9.57 & \second{69.37} & 74.11 & 83.58 \\
  & & OriAny.V2 & 28.25 & 26.02 & \second{38.76} & \second{62.36} & 10.10 & 37.31 & \second{51.47} & \second{73.99} & 9.80 & 41.95 & 57.90 & 78.04 \\
  & & Object-Uni w/o & \second{28.03} & \second{29.35} & 37.01 & 54.38 & \second{9.10} & \second{38.60} & 51.18 & 73.67 & \best{9.07} & 68.05 & \second{74.61} & \second{84.62} \\
  & & \ourscell{Object-Uni} & \best{24.13} & \best{35.23} & \best{45.79} & \best{64.66} & \best{8.15} & \best{42.10} & \best{55.89} & \best{77.50} & \second{9.14} & \best{70.02} & \best{77.23} & \best{86.37} \\
  \cmidrule(lr){2-15}
  
  & \multirow{7}{*}{\begin{sideways}Multi\end{sideways}} 
  & Random & 90.43 & 2.26 & 4.29 & 10.10 & 49.25 & 4.51 & 7.42 & 17.74 & 91.77 & 1.03 & 2.42 & 8.40 \\
  & & GPT-4o & 55.39 & 15.68 & 20.53 & 31.89 & 14.17 & 25.14 & 35.36 & 59.55 &  12.72 & 63.01 & 69.95 & \second{81.47}\\
  & & Gemini-2.5-pro & 55.00 & 11.58 & 16.83 & 30.58 & 14.49 & 17.90 & 32.26 & 60.95 & 12.80 & 62.12 & 67.86 & 78.17 \\
  & & OriAny.V1 & 52.11 & 13.77 & 22.14 & 41.88 & 24.98 & 12.74 & 19.67 & 40.83 & 12.30 & \second{65.07} & \second{70.43} & 79.96 \\
  & & OriAny.V2 & \second{34.40} & \second{22.67} & \second{35.15} & \best{59.57} & \second{10.30} & \best{36.78} & \best{51.35} & \second{73.64} & \second{10.72} & 43.90 & 59.33 & 78.51 \\
  & & Object-Uni w/o & 42.08 & 17.26 & 22.79 & 39.08 & 11.99 & 33.90 & 46.73 & 71.18 & 12.79 & 63.15 & 68.61 & 79.25 \\
  & & \ourscell{Object-Uni} & \best{33.84} & \best{24.79} & \best{35.70} & \second{56.30} & \best{9.47} & \second{36.10} & \second{50.50} & \best{74.13} & \best{9.64} & \best{66.85} & \best{72.35} & \best{82.32} \\
  \midrule
  
  \multirow{14}{*}{\begin{sideways}\textbf{Objectron}\end{sideways}} 
  & \multirow{7}{*}{\begin{sideways}Single\end{sideways}} 
  & Random & 91.66 & 1.17 & 2.35 & 7.80 & 53.95 & 2.39 & 4.71 & 15.09 & 90.15 & 1.51 & 2.89 & 8.58 \\
  & & GPT-4o & 49.50 & 7.57 & 12.98 & 30.48 & 35.16 & 2.06 & 4.26 & 16.95 & 13.01 & 32.95 & 49.31 & 73.31 \\
  & & Gemini-2.5-pro & 49.95 & 5.10 & 9.59 & 25.51 & 20.79 & 6.51 & 13.36 & 42.40 & 16.36 & 28.38 & 43.15 & 67.08 \\
  & & OriAny.V1 & 37.70 & 15.86 & 27.43 & 49.94 & 31.02 & 14.00 & 23.58 & 59.47 & 15.01 & \second{33.10} & 48.97 & 71.83 \\
  & & OriAny.V2 & 30.34 & 14.10 & 24.95 & 53.49 & 11.97 & 21.34 & 37.29 & 69.62 & 13.72 & 15.76 & 29.53 & 64.02 \\
  & & Object-Uni w/o & \second{28.90} & \second{17.60} & \second{29.28} & \second{54.63} & \second{5.98} & \second{36.19} & \second{56.40} & \second{82.59} & \second{11.73} & 32.25 & \second{49.96} & \second{75.09} \\
  & & \ourscell{Object-Uni} & \best{22.67} & \best{26.57} & \best{41.84} & \best{65.99} & \best{4.24} & \best{47.49} & \best{67.16} & \best{87.84} & \best{8.12} & \best{46.52} & \best{63.25} & \best{82.91} \\
  \cmidrule(lr){2-15}
  
  & \multirow{7}{*}{\begin{sideways}Multi\end{sideways}} 
  & Random & 90.13 & 1.25 & 2.62 & 8.26 & 58.44 & 2.38 & 4.53 & 14.86 & 92.07 & 1.32 & 2.79 & 8.75 \\
  & & GPT-4o & 69.37 & 4.12 & 6.94 & 17.79 & 45.16 & 0.23 & 0.53 & 5.72 & 23.01 & 16.43 & 27.92 & 53.00 \\
  & & Gemini-2.5-pro & 68.51 & 3.39 & 6.09 & 16.59 & 18.07 & 8.76 & 16.28 & 46.96 & 34.96 & 11.20 & 19.83 & 41.32 \\
  & & OriAny.V1 & 47.26 & 6.99 & 13.94 & 32.66 & 45.34 & 7.24 & 13.64 & 44.40 & 27.41 & 15.12 & 25.33 & 48.70 \\
  & & OriAny.V2 & \second{32.58} & 9.52 & 16.71 & \second{41.43} & 10.92 & 19.41 & 35.19 & 69.37 & 20.62 & 15.56 & 27.44 & 54.40 \\
  & & Object-Uni w/o & 33.95 & \second{10.23} & \second{18.01} & 41.27 & \second{7.42} & \second{31.21} & \second{49.36} & \second{78.16} & \second{17.71} & \second{21.82} & \second{34.65} & \second{62.29} \\
  & & \ourscell{Object-Uni} & \best{25.89} & \best{17.62} & \best{30.43} & \best{57.01} & \best{5.49} & \best{37.79} & \best{57.59} & \best{83.06} & \best{12.70} & \best{29.26} & \best{44.16} & \best{71.26} \\
  \bottomrule
  \end{tabular}%
  }
  \vspace{-2mm}
  \end{table*}

\paragraph{Settings}
We evaluate object orientation estimation on three subsets of UniSpatial-80K: KITTI-Cityscapes, ImageNet3D, and Objectron. Each dataset is evaluated under both single-object and multi-object settings. We compare our model with four baselines: GPT-4o \citep{GPT-4o}, Gemini-2.5-Pro \citep{gemini25flashimage}, Orient Anything V1~\citep{wang2025orient}, and Orient Anything V2~\citep{wangorient}. For GPT-4o and Gemini-2.5-Pro, we prompt the model with the full image and object bounding boxes, and ask it to estimate the orientation of each target object. The exact prompts are provided in the Appendix \ref{app_prompt}. For Orient Anything V1/V2, we crop the target object according to the ground-truth bounding box and feed the cropped object image into the orientation estimator.

\paragraph{Evaluation Metrics}
For orientation estimation, we report the mean angular error and Area Under the Recall Curve (AUC) under thresholds of $5^\circ$, $10^\circ$, and $30^\circ$. For azimuth and rotation, we use circular angular distance: $d(\hat{\theta}, \theta) = \min(|\hat{\theta}-\theta|, 360^\circ - |\hat{\theta}-\theta|)$. For polar angle, we use the absolute error since it is defined on $[0^\circ, 180^\circ]$.

\paragraph{Comparison Results}
Table~\ref{tab:main_orientation} reports the quantitative results. Overall, Object-Uni achieves the best performance across most datasets, splits, and orientation metrics. The gains are especially remarkable for azimuth estimation, which requires reasoning about object-facing direction and is therefore more challenging than image-plane localization. Compared with general-purpose MLLMs, Object-Uni produces more stable object-level orientation estimation, especially in multi-object scenes. It also performs competitively against specialist orientation models while remaining a unified understanding-generation model. These results demonstrate that object-centric spatial supervision and object-token grounding effectively improve orientation awareness, with the azimuth gains providing direct evidence for stronger object-centric spatial alignment.

\subsection{Pose-Controllable Generation}

\paragraph{Settings}
We evaluate object pose as a precise, instance-level control signal for image generation. Given a scene description and target object poses, the model must synthesize an image where each specified object is faithful to the prompt, correctly localized, and rendered in the desired orientation. This is harder than standard text-to-image generation because the model must bind geometric controls to the correct object instances, especially in multi-object scenes. We compare with SceneDesigner~\citep{qinscenedesigner}, a representative method for multi-object controllable generation with explicit pose manipulation. Experiments are conducted on the ImageNet3D, KITTI-Cityscapes, and Objectron subsets of our Unispatial dataset, covering general, street-view, and indoor object-centric scenes. For each dataset, we evaluate both single-object and multi-object settings.

\paragraph{Metrics}
We evaluate generated images along three axes: spatial grounding, pose controllability, and semantic fidelity. For spatial grounding, we first detect generated objects with Grounding DINO~\citep{liu2024grounding} and compare the detected boxes with the target layouts. We report mean Intersection over Union (mIoU) to measure localization overlap and spatial accuracy $Acc_{ls}=\frac{1}{N}\sum_{i=1}^{N}\mathbb{I}(\mathrm{IoU}_i > 0.6)$ to measure localization success, where $N$ denotes the number of evaluated object instances and $\mathbb{I}$ is the indicator function. For pose controllability, we estimate the orientation of each generated object with an orientation estimator~\citep{wangorient} and compute its angular deviation from the target condition. We report the mean error and AUC@10/30, which summarize the accuracy curve within $10^\circ$ and $30^\circ$ error thresholds, respectively. Finally, we report the CLIP score~\citep{radford2021learning} to measure text-image alignment. These metrics jointly assess whether the model can generate objects at the correct location, under the correct orientation, while preserving semantic consistency with the input prompt.

\begin{figure}
    \centering
    \includegraphics[width=1.0\linewidth]{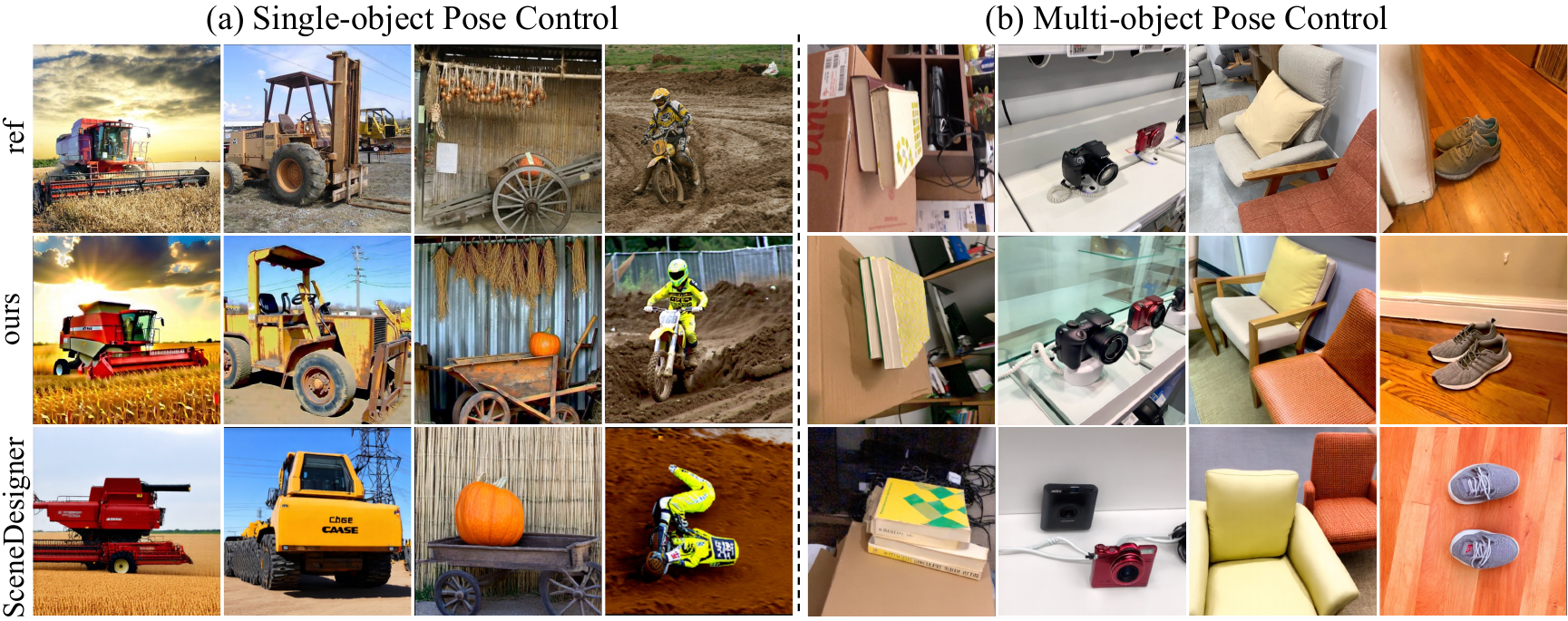}
    \caption{Evaluation of pose-controllable generation in single- and multi-object scenarios. The reference image in the first row serves as the source for the input text and 9D pose annotations. Object-Uni shows better text alignment and pose consistency under diverse pose conditions.}
    \label{fig:generation}
\end{figure}

\paragraph{Comparison Results}
As shown in Table~\ref{tab:orientation_generation}, Object-Uni achieves stronger overall performance than SceneDesigner across datasets and settings. For spatial grounding, our model achieves clear gains in both mIoU and $Acc_{ls}$, especially in multi-object scenes, indicating stronger object-level localization and layout binding. For example, on Objectron, Object-Uni improves mIoU from 40.77 to 78.23 in the single-object setting and from 27.25 to 69.08 in the multi-object setting. For pose controllability, Object-Uni generally reduces angular errors and improves AUC@10/30 for azimuth, polar, and rotation. The improvements are particularly evident on ImageNet3D and Objectron, where object appearance varies significantly with viewpoint. The advantage becomes more pronounced in multi-object scenarios, where each object must be associated with its own target orientation. Object-Uni demonstrates a stronger ability to jointly control object location, size, and orientation, while maintaining semantic consistency with the input description (see Figure~\ref{fig:generation}). This indicates that our method can better understand and exploit the input scene description, corresponding to the higher CLIP scores across all settings. More visual results are provided in Appendix \ref{app: more_results}.

\begin{table*}[t]
\centering
\scriptsize
\renewcommand{\arraystretch}{0.95}
\setlength\tabcolsep{3.4pt}
\caption{\textbf{Evaluation results on pose-controllable generation.} We compare SceneDesigner and our Object-Uni model on ImageNet3D, KITTI, and Objectron. The best results are marked in bold.}
\label{tab:orientation_generation}
\resizebox{\linewidth}{!}{
\begin{tabular}{cllcccccccccccc}
\toprule
& & & \multicolumn{2}{c}{Localization} & \multicolumn{3}{c}{Azimuth [in $^\circ$]} & \multicolumn{3}{c}{Polar [in $^\circ$]} & \multicolumn{3}{c}{Rotation [in $^\circ$]} & Gen \\
\cmidrule(lr){4-5}
\cmidrule(lr){6-8}
\cmidrule(lr){9-11}
\cmidrule(lr){12-14}
\cmidrule(lr){15-15}
 &  & Approach
& mIoU$\uparrow$
& $Acc_{ls}$$\uparrow$
& Err$\downarrow$
& \multicolumn{2}{c}{AUC@10/30$\uparrow$}
& Err$\downarrow$
& \multicolumn{2}{c}{AUC@10/30$\uparrow$}
& Err$\downarrow$
& \multicolumn{2}{c}{AUC@10/30$\uparrow$}
& CLIP$\uparrow$ \\
\midrule

\multirow{4}{*}{\begin{sideways}\textbf{ImageNet3D}\end{sideways}}
& \multirow{2}{*}{Sgl.}
& SceneD.
& 66.45 & 68.81
& 62.10 & 13.30 & 24.59
& 15.77 & 35.25 & 58.92
& 12.12 & 41.95 & 70.78
& 0.333 \\
& & \ourscell{Object-Uni}
& \textbf{74.31} & \textbf{83.40}
& \textbf{34.40} & \textbf{30.14} & \textbf{50.74}
& \textbf{10.61} & \textbf{47.32} & \textbf{71.77}
& \textbf{9.68} & \textbf{55.57} & \textbf{78.12}
& \textbf{0.341} \\
\cmidrule(lr){2-15}
& \multirow{2}{*}{Mul.}
& SceneD.
& 39.03 & 20.89
& 67.83 & 13.99 & 25.67
& 20.24 & 32.48 & 54.65
& 16.69 & 46.01 & 69.54
& 0.334 \\
& & \ourscell{Object-Uni}
& \textbf{54.43} & \textbf{44.86}
& \textbf{50.66} & \textbf{22.11} & \textbf{39.43}
& \textbf{12.64} & \textbf{42.38} & \textbf{67.36}
& \textbf{11.49} & \textbf{58.37} & \textbf{77.56}
& \textbf{0.344} \\
\midrule

\multirow{4}{*}{\begin{sideways}\textbf{KITTI}\end{sideways}}
& \multirow{2}{*}{Sgl.}
& SceneD.
& 38.40 & 26.45
& \textbf{68.17} & 10.40 & 24.60
& 4.03 & 70.79 & 87.28
& 7.86 & 53.15 & 76.30
& 0.319 \\
& & \ourscell{Object-Uni}
& \textbf{62.75} & \textbf{68.60}
& 69.83 & \textbf{12.60} & \textbf{29.59}
& \textbf{2.46} & \textbf{78.09} & \textbf{92.61}
& \textbf{4.81} & \textbf{56.96} & \textbf{84.39}
& \textbf{0.331} \\
\cmidrule(lr){2-15}
& \multirow{2}{*}{Mul.}
& SceneD.
& 17.17 & 11.72
& 89.48 & 8.28 & 20.34
& 9.00 & 83.63 & 90.28
& 20.35 & 36.17 & 55.90
& 0.315 \\
& & \ourscell{Object-Uni}
& \textbf{53.34} & \textbf{49.34}
& \textbf{81.36} & \textbf{9.63} & \textbf{29.40}
& \textbf{1.78} & \textbf{90.52} & \textbf{96.34}
& \textbf{5.96} & \textbf{54.72} & \textbf{81.69}
& \textbf{0.323} \\
\midrule

\multirow{4}{*}{\begin{sideways}\textbf{Objectron}\end{sideways}}
& \multirow{2}{*}{Sgl.}
& SceneD.
& 40.77 & 13.56
& 53.44 & 9.12 & 22.68
& 24.67 & 14.92 & 39.36
& 15.83 & 31.67 & 63.09
& 0.326 \\
& & \ourscell{Object-Uni}
& \textbf{78.23} & \textbf{93.12}
& \textbf{32.10} & \textbf{22.61} & \textbf{49.52}
& \textbf{13.63} & \textbf{32.03} & \textbf{64.98}
& \textbf{14.11} & \textbf{31.68} & \textbf{64.55}
& \textbf{0.335} \\
\cmidrule(lr){2-15}
& \multirow{2}{*}{Mul.}
& SceneD.
& 27.25 & 4.53
& 82.27 & 3.65 & 10.26
& 24.89 & 14.39 & 39.98
& 25.88 & 22.90 & 48.01
& 0.319 \\
& & \ourscell{Object-Uni}
& \textbf{69.08} & \textbf{79.37}
& \textbf{36.26} & \textbf{13.93} & \textbf{36.34}
& \textbf{11.74} & \textbf{32.48} & \textbf{67.48}
& \textbf{20.91} & \textbf{27.39} & \textbf{53.82}
& \textbf{0.334} \\
\bottomrule
\end{tabular}
}
\end{table*}

\subsection{Ablation Study}

\paragraph{Effect of Object-Token Grounding Pose Anchor}
Object-Uni introduces the proposed object tokens and binds each orientation prediction to a specific annotated object. The ablation results are included in Table~\ref{tab:main_orientation}, where Object-Uni w/o denotes the base model trained with the same tasks but without explicit object tokens for orientation prediction. The comparison shows that object-token grounding consistently improves orientation estimation, particularly for azimuth. The improvement is more pronounced in multi-object scenes, where the model must distinguish which object a predicted orientation belongs to. 

\subsection{Application: Object-Centric Novel View Synthesis}

Beyond orientation prediction and text-to-image generation, our Object-Uni model can be applied to object-centric novel view synthesis. Given an input image and a target pose, the model edits the target object so that its appearance matches the desired viewpoint while maintaining its category, local texture, and surrounding context. This task requires the model to perform spatial imagination: it must infer how the object should look after rotation rather than merely copying pixels from the input image. Visual results of object-centric novel-view synthesis are shown in Figure~\ref{fig:ability} (b). The image on the left shows the generated result; the small image in the upper-right corner shows the input image from the initial viewpoint, and the small image in the lower-right corner shows the target pose. Our model remains consistent with the input appearance while accurately matching the target pose.

\section{Conclusion}
Object-Uni is a unified framework for object-centric spatial understanding and controllable generation. Built on UniSpatial-80K with object-centric spatial supervision, it jointly models object semantics, layout, pose, and pose-conditioned generation. Experiments show strong spatial understanding and controllable generation under object-level pose constraints.

\paragraph{Limitation.}
Although Object-Uni shows strong spatial understanding and controllable generation ability, it still has limitations in generating fine-grained text and human details.
These issues are partly inherited from the underlying generative backbone and remain important directions for future improvement.

\newpage
\bibliographystyle{plain}
\bibliography{neurips_2026}

%%%%%%%%%%%%%%%%%%%%%%%%%%%%%%%%%%%%%%%%%%%%%%%%%%%%%%%%%%%%
\newpage
\appendix

\section{Orientation to viewpoint description}
\label{app: view_mapp}
To bridge the gap between continuous object poses and the abstract spatial reasoning capability learned by MLLMs, we introduce an orientation-to-language abstraction as intermediate supervision for our framework. Specifically, following the Orient-Anything convention, each object orientation is represented by three angles: azimuth, polar angle, and in-plane rotation. We quantize these continuous angles and map them to coarse semantic view descriptions: (i) Azimuth: front, front-right quarter, right side, back-right quarter, back, back-left quarter, left side, and front-left quarter; (ii) Polar: overhead, high-angle, mid-high, eye-level, mid-low, low-angle, and bottom-up; (iii) Rotation: almost no in-plane rotation and clockwise/counterclockwise rotation. As quantized abstractions of object orientation, these terms are combined with object-level spatial descriptions to express 3D object poses in a linguistically accessible form. The detailed mapping relationship between the orientation angles and semantic view terms is listed in Table~\ref{tab:orientation_language_mapping}.

\begin{table}[ht]
\centering
\small
\caption{Mapping from continuous orientation angles to viewpoint descriptions.}
\label{tab:orientation_language_mapping}
\begin{tabular}{lll}
\toprule
Angle type & Range & Language description \\
\midrule
\multirow{8}{*}{Azimuth $\phi_i$}
& $[337.5^\circ, 360^\circ) \cup [0^\circ, 22.5^\circ)$ & front \\
& $[22.5^\circ, 67.5^\circ)$ & front-right quarter \\
& $[67.5^\circ, 112.5^\circ)$ & right side \\
& $[112.5^\circ, 157.5^\circ)$ & back-right quarter \\
& $[157.5^\circ, 202.5^\circ)$ & back \\
& $[202.5^\circ, 247.5^\circ)$ & back-left quarter \\
& $[247.5^\circ, 292.5^\circ)$ & left side \\
& $[292.5^\circ, 337.5^\circ)$ & front-left quarter \\
\midrule
\multirow{7}{*}{Polar $\theta_i$}
& $[0^\circ, 15^\circ)$ & overhead \\
& $[15^\circ, 45^\circ)$ & high-angle \\
& $[45^\circ, 75^\circ)$ & mid-high \\
& $[75^\circ, 105^\circ)$ & eye-level \\
& $[105^\circ, 135^\circ)$ & mid-low \\
& $[135^\circ, 165^\circ)$ & low-angle \\
& $[165^\circ, 180^\circ]$ & bottom-up \\
\midrule
\multirow{4}{*}{Rotation ${\psi}_i$}
& $[0^\circ, 15^\circ) \cup [345^\circ, 360^\circ) $ & almost no in-plane rotation \\
& $[15^\circ, 180^\circ)$ & counterclockwise rotation \\
& $[180^\circ, 345^\circ]$ & clockwise rotation \\
\bottomrule
\end{tabular}
\end{table}

\section{Dataset details}
\subsection{Dataset statistics.}
\label{app_data}
UniSpatial-80K contains 83,252 image entries and 91,392 annotated objects from 122 categories, with an average of 1.10 objects per image.
It integrates street-view, indoor object-centric, and general rigid-object sources, covering both single-object and multi-object spatial reasoning scenarios.
The dataset exhibits a broad but moderately long-tailed category distribution: the top eight categories account for 55.16\% of object instances, while the remaining categories provide diverse object geometry and scene context.

\begin{table}[t]
\centering
\caption{Source and split composition of UniSpatial-80K. }
\label{tab:dataset_splits}
\resizebox{0.75\linewidth}{!}{
\begin{tabular}{l r r r r}
\toprule
Source & Train / Val & Test-Single & Test-Multi & File-level Total \\
\midrule
KITTI-Cityscapes & 1,266 & 363 & 303 & 1,932 \\
ImageNet3D & 39,765 & 1,229 & 146 & 41,140 \\
Objectron & 42,221 & 2,500 & 475 & 45,196 \\
\midrule
Total & 83,252 & 4,092 & 924 & 88,268 \\
\bottomrule
\end{tabular}
}
\end{table}

\begin{table}[t]
\centering
\caption{Top categories in UniSpatial-80K ranked by object count.}
\label{tab:top_categories}
\resizebox{0.85\linewidth}{!}{
\begin{tabular}{l r r r r}
\toprule
Category & Obj Count & Obj \% & Img Count & Img Coverage \\
\midrule
shoes      & 11,962 & 13.09 & 6,981 & 8.39 \\
chair      & 6,058  & 6.63  & 5,579 & 6.70 \\
bottle     & 5,744  & 6.29  & 5,726 & 6.88 \\
laptop     & 5,585  & 6.11  & 5,479 & 6.58 \\
cup        & 5,582  & 6.11  & 5,427 & 6.52 \\
camera     & 5,411  & 5.92  & 5,325 & 6.40 \\
books      & 5,050  & 5.53  & 5,004 & 6.01 \\
cereal box & 5,009  & 5.48  & 5,000 & 6.01 \\
car        & 4,096  & 4.48  & 3,688 & 4.43 \\
\bottomrule
\end{tabular}
}
\end{table}

\subsection{Prompt template}
\label{app_prompt}

The prompt used to enable general models to estimate object orientation includes definitions of both the task and the orientation representation.

\begin{tcolorbox}[
  breakable,
  colframe = gray,      
  colback = gray!5!white,             
  coltitle = white,                  
  coltext = black,                    
  fonttitle = \bfseries,             
  title = Instruction for universal orientation prediction,  
  boxrule = 1pt,                      
  arc = 2mm,                         
  width = \linewidth,               
  left = 7pt,                       
  right = 7pt,                    
  top = 5pt,                    
  bottom = 5pt                    
]

\begin{center}
\includegraphics[width=0.7\linewidth]{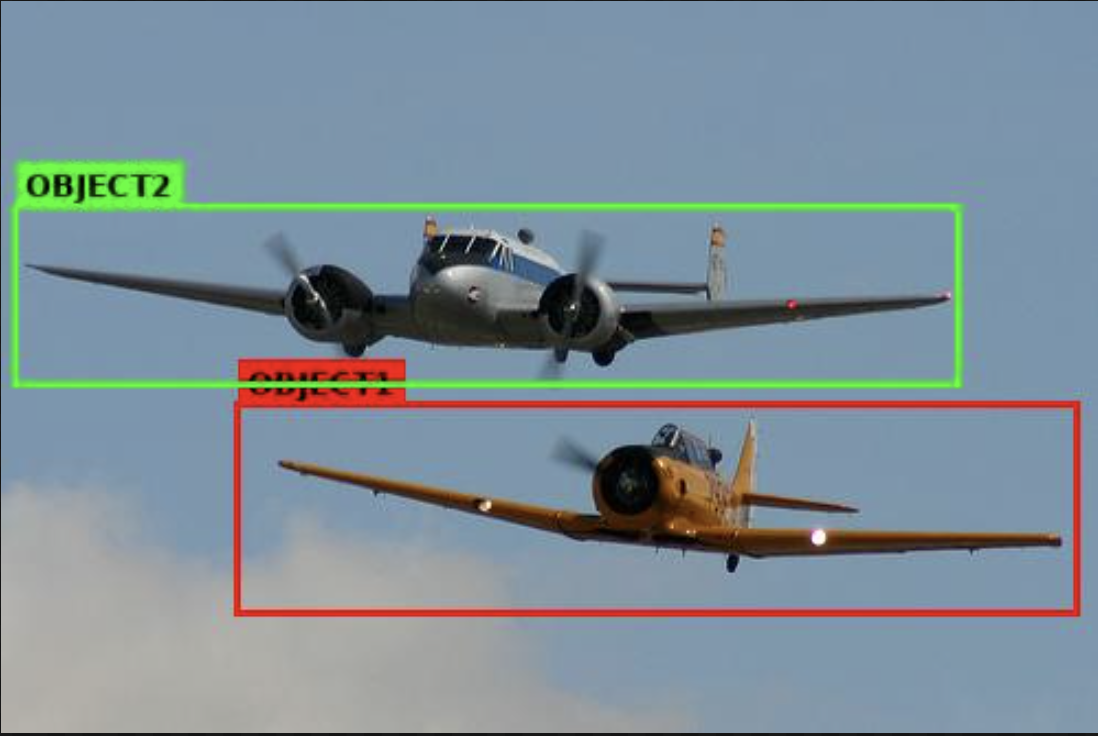}
\end{center}

\fontsize{8.5pt}{10pt}\selectfont
You are evaluating 3D object orientation in an annotated image.\\
The image contains bounding boxes and labels OBJECT1, OBJECT2, etc. Estimate orientation only for the listed objects, in exactly the listed order.\\

Objects:\\
- OBJECT1: category=\texttt{"aeroplane"}, original\_idx=1, bbox=[108, 184, 487, 277]\\
- OBJECT2: category=\texttt{"aeroplane"}, original\_idx=2, bbox=[7, 94, 433, 173]\\

Use the Orient-Anything-style angle convention:\\
1. Azimuth is the horizontal viewpoint angle around the object, in degrees within [0, 360).\\
   - 0 means front view of the object.\\
   - 90 means right-side view.\\
   - 180 means back view.\\
   - 270 means left-side view.\\
   - Front-left quarter view should be near 315 or 300; front-right quarter view should be near 45 or 60.\\
2. Polar is the vertical viewpoint angle, in degrees within [0, 180].\\
   - 0 means top/overhead view.\\
   - 90 means eye-level/horizontal view.\\
   - 180 means bottom/upward view.\\
3. Rotation is the in-plane image rotation, in degrees within [0, 360).\\
   - 0 means upright in the image plane.\\
   - 90 means clockwise rotation.\\
   - 180 means upside-down.\\
   - 270 means counter-clockwise rotation.\\

Important rules:\\
- Estimate the physical object viewpoint, not the 2D box position.\\
- Use the annotated labels in the image to identify objects.\\
- If the object is symmetric or ambiguous, still give the best single estimate.\\
- Return exactly 2 objects.\\
- Return only valid JSON. No markdown. No explanation.\\

Required JSON schema:\\
\{"objects": [\{"name": "OBJECT1", "azimuth": 0, "polar": 90, "rotation": 0\}]\}
\label{tab: general_ori}
\end{tcolorbox}

\section{Additional results}
\label{app: more_results}

\begin{figure}
    \centering
    \includegraphics[width=1\linewidth]{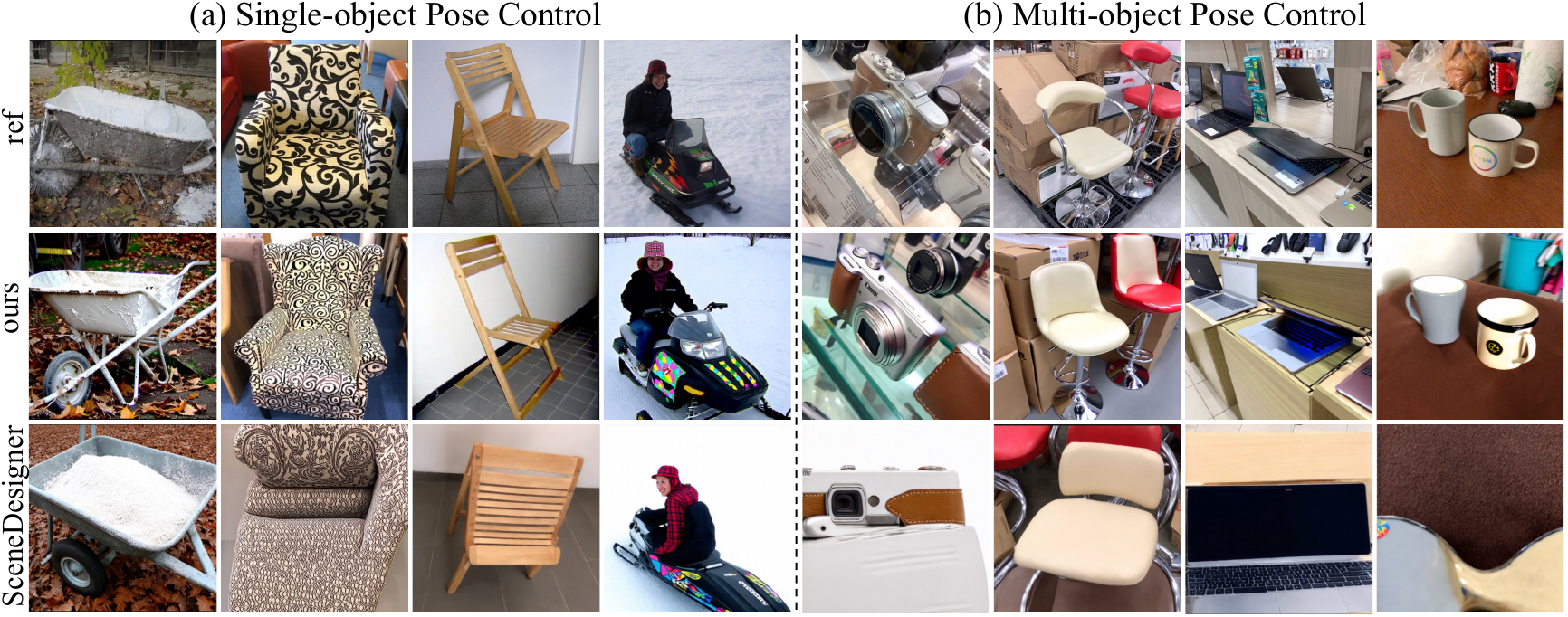}
    \caption{Additional results of pose-controllable generation in single- and multi-object scenarios. The reference image in the first row serves as the source for the input text and 9D pose annotations.}
    \label{fig:app_gen}
\end{figure}

The additional results of pose-controllable generation are shown in Figure~\ref{fig:app_gen}.

\end{document}